\documentclass[runningheads]{llncs}
\usepackage[utf8]{inputenc}
\usepackage[T1]{fontenc}
\usepackage{amsmath,amssymb}
\usepackage{graphicx}
\usepackage[
    bookmarks=false,
    hyperindex=false,
    hyperfootnotes=false,
    hypertexnames=false,
    draft=true,
    pdfborder={0 0 0},
    colorlinks=false,
    hidelinks
]{hyperref}
\usepackage{booktabs}
\usepackage{tikz}
\usetikzlibrary{shapes,arrows,positioning,fit,backgrounds}
\usepackage{algorithm}
\usepackage{algpseudocode}
\usepackage{xcolor}
\usepackage{geometry}
\begin{document}
\pagestyle{empty}
\title{Utility-Aware Data Pricing: Token-Level Quality and Empirical Training Gain for LLMs}
%
%
\author{Minghui Xu\inst{1} \and
Qi Luo\inst{1} \and
Kun Li\inst{1} \and
Zhengyang Shan\inst{1}}
\institute{Shandong University, Jinan, China \\
\email{\{mhxu,qiluo, kunli\}@sdu.edu.cn}, 202520912@mail.sdu.edu.cn \\
* Corresponding author: Minghui Xu }

\maketitle              
\begin{abstract}
Traditional ``row-count $\times$ quality coefficient'' approaches fail to capture the nonlinear utility of data for Large Language Model (LLM) capabilities. We present a utility-aware data valuation framework that moves from static accounting toward utility-based pricing. The framework combines token-level information density and data quality, empirical training contribution estimated by influence functions, proxy models, and Data Shapley values, and cryptographic mechanisms based on hash commitments, Merkle trees, and tamper-evident training ledgers. Preliminary evaluation includes a smoke-scale multi-domain study and a synthetic duplication test. Results show that proxy-based empirical gain achieves strong ranking agreement with proxy-defined realized utility and substantially outperforms row-count and token-count baselines in the tested setting, while the prototype provides a controlled sanity check of the valuation pipeline and its sensitivity to duplication-based manipulation.

\end{abstract}
\section{Introduction}
\label{sec:introduction}
The emergence of Large Language Models (LLMs) has fundamentally transformed artificial intelligence, creating unprecedented demand for high-quality training data \cite{brown2020language,touvron2023llama}. Data marketplaces have traditionally priced data by quantity, typically using ``row-count $\times$ quality coefficient'' \cite{harris2022data}, where quality coefficients capture completeness, accuracy, timeliness, and consistency. While effective for conventional analytics and classical machine learning, such models become inadequate for modern AI systems, where a high-quality reasoning chain may contribute more to model capability than millions of repetitive exchanges.

The research community has explored several directions beyond row-based pricing. For data attribution, influence functions \cite{hampel1974influence,koh2017understanding} trace the effects of training points through the loss landscape, while gradient-based methods such as TracIn \cite{pruthi2020estimating} measure gradient alignment across training checkpoints. Data Shapley values \cite{ghorbani2019data,jia2019towards} provide a game-theoretic framework for allocating value among contributors based on desirable axioms such as efficiency and symmetry. For information content, Shannon entropy \cite{shannon1948mathematical} and perplexity-based filtering are widely used to identify low-information or corrupted data. For trust and verifiability, cryptographic tools---including commitment schemes, Merkle trees, zero-knowledge proofs \cite{goldwasser1989knowledge}, and zk-SNARKs \cite{groth2016snark}---have been applied to verifiable inference and model integrity. More broadly, data-centric AI \cite{coleman2020selection} emphasizes data quality over quantity, while neural scaling laws \cite{kaplan2020scaling,hoffmann2022training} characterize how data utility changes with model scale.

Despite this progress, existing approaches primarily operate on rows, documents, or samples, which do not fully match the granularity at which modern AI systems process information. Transformer-based architectures \cite{brown2020language,rae2021scaling} learn from \emph{tokens}, creating three limitations. First, \textbf{cross-modal comparability} is limited: equal-length Python code and natural language can be compared as token sequences, whereas row-based pricing treats them as equivalent units regardless of information density. Second, \textbf{granular attribution} is obscured: although LLM training updates are computed over token positions, conventional valuation assigns a single score to an entire document. Third, \textbf{redundancy compression} is ignored: tokenization represents frequent sequences with shorter subword units, while row-based pricing treats all rows equally. These limitations motivate valuation signals that capture token-level information characteristics while assessing sources by their contribution to model training. Moreover, opaque valuation mechanisms create information asymmetry \cite{harris2022data}: sellers cannot substantiate pricing claims, buyers cannot verify whether premium data improves models, and organizations face increasing pressure for auditable data provenance.

This paper proposes a comprehensive framework for dynamic data valuation based on three foundational pillars:
\begin{enumerate}
\item \textbf{Token-level Information Density}: Moving beyond row-based counting by incorporating information entropy \cite{shannon1948mathematical}, semantic richness, and syntactic coherence at the token level.
\item \textbf{Empirical Training Gain}: Quantifying data contribution through influence functions, lightweight proxy model strategies, and Data Shapley values \cite{koh2017understanding,ghorbani2019data}, with the proxy model serving as a tractable estimator for the target LLM scale.
\item \textbf{Cryptographic Verifiability}: Providing auditable, tamper-evident records of data contribution and valuation through cryptographic mechanisms \cite{groth2016snark,goldwasser1989knowledge}.
\end{enumerate}

Our framework moves from static, quantity-based pricing toward dynamic, utility-based valuation that better reflects task-dependent data contribution. The lightweight proxy model estimates data contribution without repeatedly training large LLMs, while the cryptographic layer provides integrity and auditability for valuation records. The effectiveness of proxy-based valuation depends on the stability of relative data contributions across model scales, motivating further validation on larger target models. An open-source implementation is available at \texttt{https://github.com/BDS-SDU/utility-aware-data-pricing}.

The remainder of this paper is organized as follows. Section~\ref{sec:impact} presents our valuation framework, including token-level quality metrics, proxy-based empirical gain, influence functions, Shapley values, and the unified ensemble. Section~\ref{sec:verifiability} introduces the cryptographic verifiability layer. Section~\ref{sec:experiments} presents the preliminary experimental evaluation and reports results, and Section~\ref{sec:conclusion} concludes the paper and discusses future directions.



\section{Impact-Driven Valuation: Measuring Training Dividends}
\label{sec:impact}
 
Our valuation framework operates in four parts: token-level quality assessment (information density, syntactic coherence, and semantic richness) for fast ex-ante estimates; a lightweight proxy model and value function; three complementary training-dividend estimators (leave-one-source-out proxy gain, influence function attribution, and Monte Carlo Shapley values); and a unified ensemble score.
 
\subsection{Token-Level Quality Assessment}

We establish ex-ante quality metrics at token granularity that can be computed before any training occurs. All text is tokenized using a lightweight regex-based tokenizer that matches alphanumeric words (\texttt{[A-Za-z0-9\_]}) and individual punctuation marks from lowercased input, handling natural language, code, and mixed-format text uniformly without a pretrained subword vocabulary. These metrics characterize information and structural properties at the token level, while the resulting document-level scores are aggregated to the source level for the valuation experiments in Section~\ref{sec:experiments}. Thus, the current framework uses token-level information as a basis for quality assessment, but does not attempt to attribute training contribution to individual tokens.

\subsubsection{Information Density Metric}

We measure information density as the mean surprisal (in bits) of each token given its context, $H(X) = -\frac{1}{n} \sum_{i=1}^{n} \log_2 p(x_i | x_{<i})$, where $p(x_i | x_{<i})$ is estimated by an additive-smoothed trigram model:

\begin{equation}
\hat{p}(x_i | c) = \frac{c(c, x_i) + \alpha}{c(c) + \alpha \cdot |V|}
\end{equation}

with $c = (x_{i-2}, x_{i-1})$, $c(\cdot)$ denoting n-gram counts, and $|V|$ the vocabulary size. Unseen contexts default to the uniform distribution $1/|V|$. This reference model requires no GPU and trains in a single pass. The raw score is normalized to $[0, 1]$ by clamping at 8 bits:

\begin{equation}
\text{InfoDensity}_{\text{norm}}(X) = \min\left(\frac{H(X)}{8.0}, 1.0\right)
\end{equation}

\subsubsection{Syntactic Coherence}

Syntactic coherence measures whether token sequences conform to the structural patterns of their domains. Rather than relying on a full parser, we compute it as the unweighted average of 9 structural checks, each producing a score in $[0, 1]$:

\begin{equation}
\text{SynCo}(X) = \frac{1}{9} \sum_{k=1}^{9} c_k(X)
\end{equation}

The checks are: (1--3) balanced delimiters for \texttt{()}, \texttt{[]}, \texttt{\{\}} (binary); (4) even quote count (binary); (5) line-length sanity, $\max(0, 1 - \text{fraction of lines} > 240 \text{ chars})$; (6) character-level alphanumeric ratio; (7) token-level alphabetic ratio; (8) punctuation density penalty, $\max(0, 1 - 2 \cdot \text{punct\_token\_ratio})$; (9) malformed-token penalty, $\max(0, 1 - \text{malformed\_ratio})$, where malformed tokens include alphanumeric mixes (e.g.\ \texttt{x3f}), consonant-only strings of length $\geq 6$ (e.g.\ \texttt{bcdfgh}), and multi-character non-alphanumeric tokens.

This multi-check approach provides a lightweight, domain-agnostic structural signal across natural language, code, and mixed-format documents without requiring domain-specific configuration.

\subsubsection{Semantic Richness}

Semantic richness captures the diversity of meaning across segments of a document. The text is split into up to 16 sentence-level segments, each mapped to a 128-dimensional L2-normalized hash-based bag-of-words vector:

\begin{equation}
    e_i = \frac{\text{BoW}_{\text{hash}}(s_i)}{\|\text{BoW}_{\text{hash}}(s_i)\|_2}
\end{equation}

where $\text{BoW}_{\text{hash}}(s_i)$ accumulates token counts into a 128-dim vector indexed by $\text{SHA-256}(t) \bmod 128$. The diversity score is:

\begin{equation}
\text{diversity}(X) = 1 - \frac{2}{S(S-1)} \sum_{i < j} \text{cosine-sim}(e_i, e_j)
\end{equation}

clamped to $[0, 1]$. To downweight documents dominated by numeric noise or repetitive tokens, we apply a lexical quality factor $L(X) = 0.7 \cdot \frac{|{t \in X : t \text{ is alphabetic}}|}{|X|} + 0.3 \cdot \frac{|\text{unique}(X)|}{|X|}$, giving the final score:

\begin{equation}
\text{SemRich}(X) = \text{diversity}(X) \cdot L(X)
\end{equation}

For single-segment documents, we return a default of 0.5. The overall \textbf{Data Quality Score (DQS)} then combines the three dimensions as a weighted sum:

\begin{equation}
\text{DQS}(X) = 0.4 \cdot \text{InfoDensity}_{\text{norm}}(X) + 0.3 \cdot \text{SynCo}(X) + 0.3 \cdot \text{SemRich}(X)
\end{equation}

The weights assign the highest importance to information density as the primary information-content signal. The score is computed per document and aggregated to source-level means for use in the unified ensemble. Its effectiveness as a valuation signal is expected to depend on the downstream task and data characteristics.

\subsection{Proxy Model and Empirical Value Function}

\subsubsection{The Proxy Model}

Directly measuring data value on a full-scale LLM is prohibitively expensive. Instead, we employ a lightweight proxy model that preserves the essential structure of the valuation problem while remaining computationally tractable. The proxy model is a logistic regression classifier with hash-based text featurization. Given a training set $\mathcal{D} = \{(x_1, y_1), \ldots, (x_n, y_n)\}$ where $x_i$ is the featurized text and $y_i \in \{0, 1\}$ is the label, the model minimizes the regularized log-loss:

\begin{equation}
    \theta^* = \arg\min_{\theta} \frac{1}{n} \sum_{i=1}^{n} \mathcal{L}(z_i; \theta) + \frac{\lambda}{2} \|\theta\|^2
\end{equation}

where $\theta = (w, b)$ represents the weight vector and bias, $\mathcal{L}(z_i; \theta) = -[y_i \log \sigma(w \cdot x_i + b) + (1 - y_i) \log(1 - \sigma(w \cdot x_i + b))]$ is the per-sample log-loss, $\sigma(\cdot)$ is the sigmoid function, and $\lambda > 0$ is the L2 regularization strength. The hash-based featurization maps each token to a fixed-dimensional vector via a hash function, making the model applicable to arbitrary text without vocabulary constraints. The L2 regularization term serves two purposes: it prevents overfitting on small proxy training sets, and it ensures that the loss landscape is strongly convex, which is essential for the influence function analysis that follows.

The proxy model is used as a computationally tractable estimator of relative data contribution rather than as a replacement for the target LLM. The underlying assumption is that the relative ordering of source contributions is sufficiently stable between the proxy and the target training regime. This assumption makes proxy-based valuation practical, but it is not guaranteed across model architectures, scales, tasks, or training procedures and therefore requires validation when transferred to a specific target model.

\subsubsection{The Value Function}

To quantify model performance on a validation set $\mathcal{V}$, we define a value function $V(S)$ measuring the utility of a model trained on subset $S$:

\begin{equation} 
V(S) = \underbrace{(\bar{p}_+ - 0.5 \cdot \bar{p}_-)}_{\text{task\_utility}} - \underbrace{\frac{1}{|\mathcal{V}|} \sum_{z \in \mathcal{V}} \mathcal{L}(z; \theta_S)}_{\text{log\_loss}} \end{equation}

where $\theta_S$ are the parameters trained on $S$, $\bar{p}_+$ and $\bar{p}_-$ are the mean predicted probabilities for positive and negative validation examples respectively. The task utility term rewards discriminative ability while the log-loss term penalizes uncertainty; together they form the common value function used by the valuation methods considered below.

\subsubsection{Leave-One-Source-Out Proxy Gain}

The most direct way to measure a data source's contribution is through ablation. For a collection of data sources $\mathcal{D} = \{D_1, \ldots, D_n\}$, the proxy gain of source $D_i$ is:

\begin{equation}
G_i = V(\mathcal{D}) - V(\mathcal{D} \setminus {D_i})
\end{equation}

A positive $G_i$ indicates that $D_i$ improves the defined proxy utility, while a negative $G_i$ indicates that removing $D_i$ increases the proxy value and therefore that the source is harmful or redundant under this value function. Proxy gain serves as the backbone of our valuation pipeline and shows strong rank correlation with the corresponding proxy-defined realized utility in the preliminary evaluation in Section~\ref{sec:experiments}. Importantly, this agreement measures the internal consistency of the proxy valuation procedure rather than providing an independent validation of utility on a target LLM, since both quantities are derived from the same proxy value function.

To estimate gains at target-model scale, we apply a neural scaling law extrapolation
\begin{equation}
\hat{G}_i^{\text{target}} = G_i^{\text{proxy}} \cdot \left(\frac{N_{\text{proxy}}}{N_{\text{target}}}\right)^\alpha,
\end{equation}
where $\alpha = 0.28$ is used as an empirical default coefficient. This coefficient is intended as a practical scaling parameter rather than a universally valid constant; its transferability across model architectures, tasks, and training regimes requires further empirical validation. The same extrapolation is applied during Shapley subset evaluations to express value estimates in the target-model scale.
 
\subsection{Influence-Based Attribution}

While leave-one-source-out ablation directly measures empirical impact, it requires retraining the proxy model for each source. Influence functions offer a gradient-based alternative that approximates the effect of data points without retraining, making them useful as a complementary attribution signal at the document level and, after aggregation, at the source level.

\subsubsection{The Influence Function}

The influence of a training point $z$ on a validation set $\mathcal{V}$ quantifies how much the model's validation loss would change if $z$ were upweighted during training:

\begin{equation}
\mathcal{I}(z, \mathcal{V}) = -\nabla_\theta \bar{\mathcal{L}}(\mathcal{V}; \theta)^\top H_\theta^{-1} \nabla_\theta \mathcal{L}(z; \theta)
\end{equation}

The training gradient $\nabla_\theta \mathcal{L}(z; \theta)$ captures the direction $z$ pushes the parameters. For our proxy model, the gradient of the log-loss with respect to the weight vector is $\nabla_w \mathcal{L} = (\sigma(w \cdot x + b) - y) \cdot x$ and with respect to the bias is $\nabla_b \mathcal{L} = \sigma(w \cdot x + b) - y$.

The inverse Hessian-vector product $H_\theta^{-1}$ transforms it into curvature space; and the dot product with the validation gradient measures alignment with the direction that improves validation performance. For the regularized logistic loss, the Hessian-vector product is:

\begin{equation}
H_\theta v = \lambda v + \frac{1}{n}\sum_{i=1}^{n} \sigma(\theta \cdot x_i)\big(1 - \sigma(\theta \cdot x_i)\big) \cdot (v \cdot x_i) \cdot x_i
\end{equation}

We compute $H_\theta^{-1} v$ via the conjugate gradient (CG) method, which iterates without forming $H_\theta^{-1}$ explicitly---each iteration requires only one Hessian-vector product. In our implementation, CG runs for up to 40 iterations with tolerance $10^{-6}$, typically converging in far fewer steps due to the well-conditioned Hessian from L2 regularization.

\subsubsection{Source-Level Influence}

To compute the influence score for an entire data source $D_i$ (rather than a single document), we average the influence of its constituent documents:

\begin{equation}
\mathcal{I}_{\text{source}}(D_i, \mathcal{V}) = \frac{1}{|D_i|} \sum_{z \in D_i} \mathcal{I}(z, \mathcal{V})
\end{equation}

In practice, we compute this over up to 5 representative documents per source to balance accuracy with computational cost. The source-level influence provides an attribution signal that complements the coarser leave-one-source-out proxy gain. Influence-based scores are also more sensitive to the quality of the Hessian approximation than proxy gains because they depend explicitly on local curvature information.

In our preliminary smoke-scale experiments, influence scores exhibit negative rank correlation with the corresponding proxy-defined realized utility in several domains. This behavior may reflect the sensitivity of influence estimates to local loss curvature and Hessian approximation when the training set is extremely small. We therefore treat influence as a complementary valuation signal rather than a standalone indicator of data value in the current setting. More systematic validation with larger datasets and direct comparisons with established attribution methods such as TracIn are left for future work.
 
\subsection{Data Shapley Values}
 
For scenarios requiring fair allocation of value among multiple data contributors, we turn to cooperative game theory. Shapley values \cite{ghorbani2019data,jia2019towards} provide a principled value allocation satisfying four desirable axioms: efficiency (total value is fully distributed), symmetry (equivalent contributors receive equal value), dummy player (non-contributors receive nothing), and additivity (values are additive across tasks). For a collection of $n$ data sources $\mathcal{D} = \{D_1, \ldots, D_n\}$ and value function $V(\cdot)$, the Shapley value of source $i$ is:
 
\begin{equation}
    \phi_i = \sum_{S \subseteq \mathcal{D} \setminus \{D_i\}} \frac{|S|!(n - |S| - 1)!}{n!} \left[ V(S \cup \{D_i\}) - V(S) \right]
\end{equation}
 
The combinatorial weight ensures each ordering is equally likely, so $\phi_i$ reflects the average marginal contribution of $D_i$ across all possible coalitions.
 
\subsubsection{Monte Carlo Estimation}
 
Exact Shapley computation requires $2^n$ subset evaluations. We use Monte Carlo estimation via random permutation sampling:
 
\begin{equation}
    \hat{\phi}_i = \frac{1}{m} \sum_{j=1}^{m} \left[ V(S_j^{(i)} \cup \{D_i\}) - V(S_j^{(i)}) \right]
\end{equation}
 
where $S_j^{(i)}$ is the set of sources preceding $D_i$ in the $j$-th random permutation. Each permutation produces one marginal contribution sample for every source simultaneously. We use $m = 64$ permutations with a subset evaluation cache to avoid redundant proxy-model retraining.
 
\subsection{Unified Valuation Framework}

The valuation methods capture complementary aspects of data value: proxy gain measures empirical impact, influence functions provide gradient-based attribution, and Shapley values provide a principled allocation mechanism. Together with the ex-ante DQS, we combine these four signals into a unified score:

\begin{equation}
\hat{V}_{\text{unified}}(D_i) = w_1 \cdot \text{DQS}_{\text{norm}}(D_i) + w_2 \cdot \hat{G}_{\text{norm}}(D_i) + w_3 \cdot \mathcal{I}_{\text{norm}}(D_i) + w_4 \cdot \phi_{\text{norm}}(D_i)
\end{equation}

where each component is min-max normalized to $[0, 1]$: $x_{\text{norm}}(D_i) = \frac{x(D_i) - \min_j x(D_j)}{\max_j x(D_j) - \min_j x(D_j)}$, and the weights $w_1, w_2, w_3, w_4$ sum to 1 with defaults $(0.25, 0.35, 0.20, 0.20)$. These weights can alternatively be calibrated on a held-out validation split. The relative benefit of different weighting schemes is task- and data-dependent, and the preliminary evaluation in Section~\ref{sec:experiments} illustrates this dependence.

\subsubsection{Confidence Intervals}
 
Each valuation method provides a point estimate with inherent uncertainty.
As a simple heuristic for risk-aware pricing, we summarize the dispersion
among the three empirical valuation signals (proxy gain, influence score,
and Shapley value) using:
 
\begin{equation}
    \text{CI}_{0.95}(D_i) = \left[\bar{V}_i - 1.96 \cdot \frac{\sigma_i}{\sqrt{k}}, \;\bar{V}_i + 1.96 \cdot \frac{\sigma_i}{\sqrt{k}}\right]
\end{equation}
 
where $\bar V_i$ is the mean of the three empirical valuation signals and $\sigma_i$ is their sample standard deviation. We use this interval as a heuristic uncertainty indicator rather than a statistically calibrated
confidence interval. A source with a high mean but large dispersion across valuation methods indicates greater valuation uncertainty, whereas a narrow interval suggests stronger agreement among the estimators.
 
\subsubsection{From Valuation to Pricing}
 
The unified score $\hat{V}_{\text{unified}}$ measures relative data quality but does not directly prescribe a monetary price. We define a pricing function combining a volume-based baseline with a quality-driven premium:
 
\begin{equation}
    P(D_i) = p_{\text{base}} \cdot |D_i|_{\text{tokens}} \cdot \big(1 + \beta \cdot \hat{V}_{\text{unified}}(D_i)\big)
\end{equation}
 
where $p_{\text{base}}$ is the base price per token, $|D_i|_{\text{tokens}}$ is the token count, and $\beta \geq 0$ controls quality differentiation. This formulation is volume-sensitive (larger sources earn more), quality-differentiating (equal-size sources can be priced differently), and non-negative (since $\hat{V}_{\text{unified}} \in [0,1]$). When $\beta = 0$ it degenerates to pure token-count pricing. Confidence intervals from the previous section propagate naturally into price ranges, enabling risk-aware bidding without a centralized auction mechanism.



\section{The Verifiability Layer: Establishing Trust in Value}
\label{sec:verifiability}

The valuation methods described in previous sections provide mechanisms for estimating data value, but they do not inherently provide assurance that the resulting records have not been manipulated. This section introduces a cryptographic verifiability layer built on two primitives---hash-based commitments and Merkle trees---and a hash-chained training ledger. The resulting mechanism provides tamper-evident records and supports post-hoc auditing without requiring the full complexity of zero-knowledge proof systems. Importantly, the current layer establishes record integrity rather than proving the computational correctness of the underlying training or valuation procedure.
 
\subsubsection{Hash-Based Commitment}
 
A commitment scheme allows a party to publish a cryptographic digest without revealing the underlying value, with the guarantee that the digest cannot be ``opened'' to a different value. We use SHA-256 with a random nonce $r$; applied to model parameters $w$ and $b$, this gives:
 
\begin{equation}
    C_\theta = \text{Commit}(\theta; r) = \text{SHA-256}\big(\text{JSON}(\{w, b, r\})\big)
\end{equation}
 
where JSON serialization uses deterministic key ordering and weights are rounded to 8 decimal places to avoid floating-point nondeterminism. The nonce provides the hiding component of the commitment, while SHA-256 collision resistance provides binding under the standard cryptographic assumption.
 
\subsubsection{Merkle Tree for Data Integrity}
 
To commit to an entire dataset without revealing individual documents, we construct a Merkle tree. Given a dataset $\mathcal{D} = \{d_1, \ldots, d_n\}$, the tree is built bottom-up:
 
\begin{equation}
    \text{leaf}_i = \text{SHA-256}(d_i), \quad \text{node}_{i,j} = \text{SHA-256}(\text{node}_{i,2j-1} \| \text{node}_{i,2j})
\end{equation}
 
If the number of nodes at any level is odd, the last node is duplicated to form a pair. The root of the tree $C_\mathcal{D} = \text{root}$ serves as a compact, fixed-size commitment to the entire dataset. Two key properties make Merkle trees useful for data valuation:
 
\textbf{(1) Binding}: Any change to a document $d_i$ changes its leaf hash, which propagates up to change the root. Therefore, the root commitment $C_\mathcal{D}$ cryptographically binds the trainer to the exact dataset used.
 
\textbf{(2) Efficient verification}: To prove that a specific document $d_i$ was part of the committed dataset, one only needs to provide the leaf hash and the $\log_2 n$ sibling hashes along the path from leaf to root---an $O(\log n)$ proof rather than requiring the entire dataset.
 
\subsection{Proof of Training Ledger}
 
The verifiability layer also maintains a training ledger that records an audit trail linking the committed input data to the trained model parameters. The ledger does not attempt to prove the correctness of each gradient step, which would require zk-SNARKs or similar computational-verification machinery. Instead, it provides a tamper-evident record that can be checked after the training process.
 
\subsubsection{Ledger Construction}
 
The ledger is constructed in three stages, corresponding to the training lifecycle:

\textbf{Stage 1 (Initialization)}: Given the list of training document IDs $\{id_1, \ldots, id_n\}$, compute the dataset commitment as the Merkle root $C_\mathcal{D} = \text{MerkleRoot}(id_1, \ldots, id_n)$. Then record the initial model parameter commitment $C_0 = \text{Commit}(\theta_0)$, where $\theta_0$ represents the randomly initialized parameters.

\textbf{Stage 2 (Training)}: After each training step $t$ (or at designated checkpoints), record a ledger entry containing the step number, the batch of document IDs used in that step, the evaluation metrics on the validation set, and the parameter commitment $C_t = \text{Commit}(\theta_t)$ for the updated parameters.

\textbf{Stage 3 (Fingerprint)}: Generate a hash chain linking all entries together to provide tamper evidence:
 
\begin{equation}
    h_0 = C_0, \quad h_t = \text{SHA-256}(h_{t-1} \| \text{JSON}(\text{entry}_t))
\end{equation}
 
Each hash in the chain depends on all previous entries, so modifying any single entry invalidates every subsequent hash. The final fingerprint includes the dataset root $C_\mathcal{D}$, the initial and final parameter commitments ($C_0$ and $C_T$), the number of recorded entries, and the tail of the hash chain $h_T$.
 
\subsubsection{Ledger Verification}
 
An auditor verifies integrity by checking four properties: (1)~recomputing $C_\mathcal{D}$ from the document IDs and confirming it matches the committed root; (2)~replaying the hash chain from $C_0$ through all entries and verifying the final hash matches $h_T$; (3)~confirming that $C_0$ matches the expected initial parameters and $C_T$ matches the claimed trained model; and (4)~checking that the recorded evaluation metrics and valuation results are internally consistent with the ledger entries. These checks establish the integrity and traceability of the recorded evidence, but do not independently verify that the underlying training or valuation computation was executed correctly.
 
\subsection{Extension Paths}
The current hash-based ledger provides tamper-evident auditing. Two natural extensions could further strengthen verifiability and auditability, but are left for future work.
 
\subsubsection{Zero-Knowledge Proofs}
Zero-knowledge proofs (ZKPs) could be layered on the current commitments and Merkle tree to allow sellers to prove valuation properties, such as $\mathrm{DQS} \geq \tau$ or a claimed contribution $\hat{V}$, without revealing the underlying data. A zk-SNARK could provide such proofs, although encoding the valuation process as an arithmetic circuit would introduce additional implementation complexity.
 
\subsubsection{On-Chain Anchoring}
For third-party auditability and regulatory compliance, the training fingerprint $(C_\mathcal{D}, C_0, C_T, h_T)$ and a timestamp could be anchored on a blockchain as an additional immutable audit record. Smart contracts could further maintain seller reputation based on the historical consistency of valuation results. These extensions are beyond the scope of the current prototype.



\section{Experimental Validation Protocol}
\label{sec:experiments}

This section presents a preliminary empirical evaluation of the proposed valuation pipeline. The evaluation focuses on three aspects: the agreement between different valuation signals and proxy-defined realized utility, the variation of valuation behavior across task domains, and the sensitivity of the pipeline to simple volume-padding attacks. All experiments in this section use a smoke-scale configuration with a small number of training examples and only four source shards per domain; unless otherwise noted, every reported correlation, ranking outcome, and calibrated weight throughout Section~\ref{sec:experiments} should therefore be read as descriptive, proxy-based evidence rather than a statistically conclusive or generalizable validation at larger model or dataset scales.

\subsection{Experimental Protocol} 
\textbf{Corpora and model protocol.}
We evaluate the proposed pipeline on three real public fine-tuning datasets covering distinct task domains: Alpaca for general instruction following, GSM8K for mathematical reasoning, and CodeXGLUE-Python for code summarization. Each domain is partitioned into source-level contributors $D_1,\ldots,D_n$, reflecting the seller- or dataset-level granularity at which data pricing decisions are typically made. A held-out validation set is constructed for each target domain, including both in-domain examples and a small number of out-of-domain negative examples from the other domains.

The current evaluation is conducted in a smoke-test setting using 12 training examples and 6 in-domain validation examples per domain. Each domain is further divided into four source shards, and 3 out-of-domain negative validation examples are added from each competing domain, resulting in 12 validation examples for each target domain. The valuation pipeline first computes token-level information density and DQS, then estimates source-level contribution using proxy gain, influence functions, and Monte Carlo Shapley values, and finally combines these signals into the unified valuation score. The lightweight proxy model is a hash-based logistic regression classifier with 256-dimensional features. Although the overall framework is designed to support downstream validation at a larger target-model scale, the results reported in this section use the proxy value function as the realized-utility reference. This setting allows the behavior of the valuation pipeline to be examined at low computational cost while making the scope of the current evidence explicit.
 
\textbf{Baselines and metrics.}
We compare eight valuation signals: row count, token count, DQS-only, proxy gain, influence-only, Shapley-only, unified valuation with fixed weights, and unified valuation with calibrated weights. Row-count and token-count methods represent quantity-based pricing baselines, while the four single-component methods isolate the individual valuation signals introduced in Section~\ref{sec:impact}. 

The fixed-weight unified score uses the default weights $(0.25, 0.35, 0.20, 0.20)$ for DQS, proxy gain, influence, and Shapley, respectively. The calibrated variant uses validation evidence to adjust the relative weights of these valuation signals. The comparison therefore examines both manually specified and task-dependent combinations of the proposed components. 
 
Source-level ranking quality is assessed against the proxy-defined leave-one-source-out gain computed by the valuation pipeline. We use Spearman rank correlation as the primary measure of agreement between each valuation signal and this proxy-defined reference, with Top-$k$ overlap reported as a complementary ranking metric for identifying high-value source contributors; given the small number of shards per domain, these metrics are reported descriptively rather than accompanied by significance claims.

\subsection{Real Multi-Domain Smoke Results}
 
We first examine the behavior of the complete valuation pipeline on the three real public datasets. The deliberately small configuration provides a controlled smoke test of whether the proposed signals produce meaningful source-level rankings across heterogeneous domains. Table~\ref{tab:smoke_config} summarizes the experimental configuration.
 
\begin{table}[!ht]
\centering
\begin{tabular}{lccc}
\toprule
Dataset & Domain & Train & Val \\
\midrule
Alpaca & Instruction following & 12 & 6 \\
GSM8K & Math reasoning & 12 & 6 \\
CodeXGLUE-Python & Code summarization & 12 & 6 \\
\bottomrule
\end{tabular}
\caption{Real-data smoke benchmark configuration.}
\label{tab:smoke_config}
\end{table}
 
\subsubsection{Aggregate Ranking Quality}
 
Table~\ref{tab:aggregate_ranking} reports mean Spearman $\rho$ and Top-2 overlap across the three target domains. Proxy gain achieves the strongest aggregate agreement ($\rho=0.986$) with perfect Top-2 overlap, while row-count and token-count baselines are substantially weaker ($\rho=0.203$ and $-0.007$). DQS-only and Shapley-only provide limited ranking signals, whereas influence-only shows a strongly negative correlation ($\rho=-0.932$).

The unified scores do not outperform proxy gain: the fixed-weight and calibrated ensembles achieve $\rho=0.483$ and $0.431$, respectively. This suggests that combining heterogeneous signals does not necessarily improve ranking when their behavior differs across domains. The strong performance of proxy gain indicates that direct leave-one-source-out measurement is the most informative signal under the current proxy setting.
 
\begin{table}[!ht]
\centering
\small
\begin{tabular}{lcc}
\toprule
Method & Mean Spearman $\rho$ & Mean Top-2 overlap \\
\midrule
Row count & 0.203 & 0.167 \\
Token count & -0.007 & 0.167 \\
DQS only & 0.105 & 0.000 \\
Proxy gain & \textbf{0.986} & \textbf{1.000} \\
Influence only & -0.932 & 0.000 \\
Shapley only & -0.042 & 0.167 \\
Unified (fixed weights) & 0.483 & 0.000 \\
Unified (calibrated) & 0.431 & 0.167 \\
\bottomrule
\end{tabular}
\caption{Aggregate ranking quality on the three-domain real-data smoke run.}
\label{tab:aggregate_ranking}
\end{table}
 
\IfFileExists{figures/real_summary_metrics.pdf}{
\begin{figure}[!ht]
\centering
\includegraphics[width=0.65\linewidth]{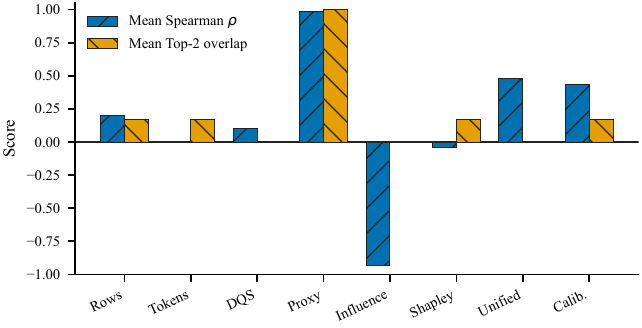}
\caption{Aggregate ranking quality in the real-data smoke benchmark.}
\label{fig:summary_metrics}
\end{figure}
}{}
 
\subsubsection{Per-Domain Breakdown}
The aggregate results conceal substantial domain-dependent variation. Table~\ref{tab:per_domain_rho} reports Spearman correlations for DQS, proxy gain, fixed-weight unified valuation, and calibrated valuation across the three target domains. Proxy gain remains the strongest signal in all domains, with $\rho=0.979$, $1.000$, and $0.979$ for code summarization, instruction following, and mathematical reasoning, respectively.

Other signals exhibit stronger domain dependence. DQS correlates positively with instruction following ($\rho=0.622$) and code summarization ($\rho=0.224$), but negatively with mathematical reasoning ($\rho=-0.531$). This likely reflects the current DQS emphasis on information density, syntactic coherence, and semantic richness, which may not directly capture the value of multi-step mathematical reasoning. Accordingly, the fixed-weight unified score achieves $\rho=0.727$ for instruction following but only $0.168$ for mathematical reasoning. Calibration improves the latter to $\rho=0.448$ by reducing the relative contributions of DQS and influence while increasing those of proxy gain and Shapley value.
 
 
\begin{table}[!ht]
\centering
\begin{tabular}{lcccc}
\toprule
Target domain & DQS & Proxy & Unified & Calibrated \\
\midrule
Code summarization & 0.224 & \textbf{0.979} & 0.552 & 0.224 \\
General instruction & 0.622 & \textbf{1.000} & 0.727 & 0.622 \\
Math reasoning & -0.531 & \textbf{0.979} & 0.168 & 0.448 \\
\bottomrule
\end{tabular}
\caption{Per-domain Spearman correlation with proxy-defined realized gain.}
\label{tab:per_domain_rho}
\end{table} 

Table~\ref{tab:complete_correlations} provides the complete correlation matrix. Influence-only scores are negatively correlated with proxy-defined realized gain in all three domains ($\rho\in\{-0.937,-0.979,-0.881\}$). Rather than interpreting this result as evidence that influence-based valuation is intrinsically unsuitable, we attribute it cautiously to the sensitivity of influence estimates to the local Hessian approximation and the extremely small sample size in the current smoke test. This result is also consistent with the discussion in Section~\ref{sec:impact}: influence functions provide a computationally efficient attribution signal, but their reliability depends on the quality of the curvature approximation. More robust gradient-based attribution methods, such as TracIn, provide a natural direction for future comparison.

The row-count baseline also varies substantially across domains, ranging from $\rho=0.888$ for instruction following to $\rho=-0.741$ for mathematical reasoning. This variation further suggests that raw document count is not a stable proxy for source contribution across heterogeneous tasks.
 
\begin{table}[!ht]
\centering
\footnotesize
\begin{tabular}{lrrrrrrrr}
\toprule
Target & Rows & Tokens & DQS & Proxy & Infl. & Shapley & Unified & Calib. \\
\midrule
Code & 0.462 & 0.119 & 0.224 & \textbf{0.979} & -0.937 & -0.259 & 0.552 & 0.224 \\
Instr. & 0.888 & -0.336 & 0.622 & \textbf{1.000} & -0.979 & -0.315 & 0.727 & 0.622 \\
Math & -0.741 & 0.196 & -0.531 & \textbf{0.979} & -0.881 & 0.448 & 0.168 & 0.448 \\
\bottomrule
\end{tabular}
\caption{Complete per-domain Spearman correlations for all estimators.}
\label{tab:complete_correlations}
\end{table}

\subsubsection{Component Weight Analysis}
The calibrated ensemble provides additional evidence that the relative usefulness of valuation signals depends on the target domain. For code summarization, the calibrated model assigns a weight of $0.814$ to proxy gain and $0.186$ to DQS. For instruction following, the corresponding weights are $0.616$ for proxy gain and $0.384$ for DQS. For mathematical reasoning, DQS and influence receive zero weight, while proxy gain and Shapley receive weights of $0.686$ and $0.314$, respectively.

These results suggest that a single manually specified mixture is unlikely to be optimal across heterogeneous tasks. At the same time, the calibrated ensemble should be viewed only as a preliminary demonstration of task-dependent weighting: the current evaluation does not establish a general calibration procedure or its stability across larger datasets. The main implication is that valuation components can provide complementary signals, but their relative importance should be determined from task- and data-specific validation evidence.

Table~\ref{tab:top_source} reports the highest-ranked source shard for each target domain. The selected source is domain-coherent in all three cases: CodeXGLUE shard~02 for code summarization, Alpaca shard~00 for instruction following, and GSM8K shard~01 for mathematical reasoning. This observation provides an additional qualitative sanity check that the valuation pipeline can distinguish source domains in the current smoke setting. The full source-by-estimator matrix is released as CSV to provide machine-readable audit evidence.

\begin{table}[!ht]
\centering
\footnotesize
\begin{tabular}{llllrrr}
\toprule
Target & Top source & Source domain & $G_i$ & Proxy & Unified & Calib. \\
\midrule
Code & CodeXGLUE shard 02 & Code & -0.013079 & -0.008956 & 0.584 & 0.140 \\
Instr. & Alpaca shard 00 & Instruction & -0.013133 & -0.008984 & 0.594 & 0.317 \\
Math & GSM8K shard 01 & Math & -0.013118 & -0.008999 & 0.523 & -0.011 \\
\bottomrule
\end{tabular}
\caption{Highest proxy-defined gain source shard for each target domain.}
\label{tab:top_source}
\end{table}
 
 
\subsection{Prototype Sanity Check}
\label{sec:prototype}
We additionally evaluate the end-to-end pipeline on a synthetic five-source benchmark containing high-quality API documentation, reasoning traces, mixed FAQ content, redundant boilerplate, and noisy text, serving as a controlled sanity check for the relative behavior of the valuation signals.

As shown in Table~\ref{tab:prototype_results}, proxy gain achieves the strongest ranking agreement with the proxy-defined realized gain ($\rho=0.90$, Top-2 overlap $=1.00$), compared with document count ($\rho=0.20$) and token count ($\rho=0.00$). The calibrated ensemble reaches $\rho=0.70$, assigning a weight of $0.69$ to proxy gain and $0.31$ to DQS, while influence and Shapley receive zero weight in this small prototype. These observations are consistent with the real-data smoke results in suggesting that proxy gain is the most informative signal under the current experimental configuration, while the optimal combination of components can depend on the data setting.
 
We further perform a simple duplication test in which noisy text is duplicated to increase its apparent volume. Under this volume-padding condition, quantity-based baselines promote the noisy source in the ranking, whereas the unified valuation retains the original top-ranked source. This provides preliminary evidence that incorporating utility-oriented signals can reduce sensitivity to simple duplication-based manipulation, though more comprehensive adversarial evaluation is left for future work.

\begin{table}[!ht]
\centering
\begin{tabular}{lcc}
\toprule
Method & Spearman $\rho$ with proxy-defined gain & Top-2 overlap \\
\midrule
Row count & 0.20 & 0.50 \\
Token count & 0.00 & 0.50 \\
DQS only & 0.40 & 0.50 \\
Proxy gain only & 0.90 & 1.00 \\
Unified (fixed weights) & 0.10 & 0.00 \\
Unified (calibrated weights) & 0.70 & 0.50 \\
\bottomrule
\end{tabular}
\caption{Prototype benchmark results against proxy-defined leave-one-source-out gain.}
\label{tab:prototype_results}
\end{table}



\section{Conclusion}
\label{sec:conclusion}
We presented a utility-aware data valuation framework for source-level data pricing in LLM-oriented applications, integrating token-level information density, empirical training contribution, and cryptographically protected audit records through DQS, proxy gain, influence functions, and Data Shapley values.

Preliminary experiments show that proxy gain provides the strongest valuation signal under the proxy-defined value function, while other signals vary across tasks. Synthetic duplication tests further suggest reduced sensitivity to volume-padding manipulation. The cryptographic layer enables tamper-evident and traceable records but does not guarantee computational correctness. Future work will evaluate the framework on larger models and datasets and develop stronger attribution and verifiability mechanisms.

\section*{Acknowledgment}
This study was supported by the Key R\&D Program of Shandong Province (No. 2025CXPT033), the National Natural Science Foundation of China (No. 62402287, U25A20425, 62302266, 62232010, U23A20302, U24A20244), the Shandong Provincial Natural Science Foundation (No.ZR2024QF214), the Research Project of Quancheng Laboratory, China under Grant No. QCL20250106, 2025 Shandong Cyberspace Administration Reform Pilot Projects, and Taishan Scholars Program under Grant tsqn202507069. 

\bibliographystyle{splncs04}
\bibliography{references}

\end{document}